\documentclass[runningheads]{llncs}
\usepackage{graphicx}
\usepackage{algorithm}
\usepackage{algorithmic}
\usepackage{pifont}
\begin{document}
\title{FedPH: Privacy-enhanced Heterogeneous Federated Learning \thanks{Supported by the National Natural Science Foundation of P.R.China under Grants [61903053], [62273065]; The Science and Technology Research Program of Chongqing Municipal Education Commission under Grants [KJZD-K201800701], [KJCX2020033]; The Opening Project of Shanghai Key Laboratory of Integrated Administration Technologies for Information Security under Grants [AGK2020006].}}
%
%
\author{Kuang Hangdong \and Mi Bo}
\authorrunning{F. Author et al.}
%
\institute{College of Information Science and Engineering, Chongqing Jiaotong University, Chongqing 400074, P.~R.~China
\email{khd401208163@gmail.com}
}

\maketitle              
\begin{abstract}
Federated Learning is a distributed machine-learning environment that allows clients to learn collaboratively without sharing private data. This is accomplished by exchanging parameters. However, the differences in data distributions and computing resources among clients make related studies difficult. To address these heterogeneous problems, we propose a novel Federated Learning method. Our method utilizes a pre-trained model as the backbone of the local model, with fully connected layers comprising the head. The backbone extracts features for the head, and the embedding vector of classes is shared between clients to improve the head and enhance the performance of the local model.
By sharing the embedding vector of classes instead of gradient-based parameters, clients can better adapt to private data, and communication between the server and clients is more effective. To protect privacy, we propose a privacy-preserving hybrid method that adds noise to the embedding vector of classes. This method has a minimal effect on the performance of the local model when differential privacy is met. We conduct a comprehensive evaluation of our approach on a self-built vehicle dataset, comparing it with other Federated Learning methods under non-independent identically distributed(Non-IID).
\end{abstract}
\section{Introduction}
Data is the fuel that powers machine learning. However, in the real world, data is often distributed across various locations, making it impossible to send private data to a central server for model training due to personal privacy concerns and data protection laws~\cite{Voigt}.

To address these challenges, the concept of Federated Learning was introduced~\cite{McMahan}, where multiple clients perform machine learning tasks with the help of a central server. Private data is kept local and is never exchanged or transferred. Federated Learning involves server aggregation and parameter updates~\cite{Kairouz} and has been successfully applied in various domains such as healthcare~\cite{W. de Brouwer}, mobile internet~\cite{Apple,Chen}, and finance~\cite{WeBank}.


The distribution of private data among different clients may result in non-independent and identically distributed(Non-IID), leading to data heterogeneity. Federated Learning researchers face a challenge in ensuring that the local model performs well when the local objective is far from the global objective, as the gradient-based aggregation method may not be effective in the presence of data heterogeneity~\cite{Kairouz}. Various studies have attempted to address this issue, such as FedProx~\cite{Li}, which limits local updates based on the $L_2$ distance between the local and global models, and FedDyn~\cite{Durmus}, which proposes a dynamic regularizer for each client at each round. However, experiments show that these methods are not effective for Non-IID datasets, as most Federated Learning methods update parameters synchronously based on the gradient space, without considering the possibility that the global model may not perform well with Non-IID datasets. Therefore, Personalized Federated Learning that personalizes the local model is crucial~\cite{Kulkarni}. Personalized Federated Learning introduces a new paradigm for collaborative learning by sharing feature embedding vectors.


Model heterogeneity is a significant challenge in Federated Learning due to the inconsistency in local model structures caused by differences in clients' computing resources. However, existing methods are not designed to handle such heterogeneity, as they rely on local model consistency for aggregation. To address this challenge, some researchers, such as Arivazhagan et al.~\cite{Arivazhagan}, proposes using a personalization layer for local models, while others, like Sattler et al.~\cite{Sattler}, suggest creating different models for various user groups. However, these approaches may not sufficiently account for data heterogeneity, particularly when the private data exists feature-shifted. To overcome model heterogeneity, it is possible to leverage the sharing of feature-embedding information in addition to data heterogeneity.


Although sharing embedding information is a common method to address model heterogeneity in Federated Learning, it may not provide sufficient data privacy assurances~\cite{Nasr}. Local differential privacy has been integrated with Federated Learning to classify images and analyze natural language~\cite{Bhowmick}, but reducing the privacy budget does not guarantee improved model performance. To resolve these issues, Sun et al.~\cite{Sun} proposed adding noise to parameters based on their value range, although the gradient explosion issue may occur during backpropagation with fewer clients. In contrast to local differential privacy, we propose a novel privacy-preserving approach that minimizes impact on the local model, ensuring that private data remains local and using multi-key semi-homomorphic encryption and differential privacy to protect data privacy.

These are our primary contributions, in brief:

1. We propose FedPH, an approach that effectively addresses the heterogeneity issue and significantly reduces communication costs by utilizing the pre-trained model as the backbone of the local model and adopting an aggregation approach to communicate embedding information.

2. We propose a novel privacy protection strategy that minimizes the impact on local model performance while ensuring differential privacy.

3. We create a vehicle dataset that considers the influence of diverse weather conditions on vehicle classification. Our results demonstrate that FedPH outperforms baseline methods.

\section{Related Work}
\subsection{Federated Learning}

McMahan et al.~\cite{McMahan} introduced FedAvg, which is a Federated Learning method that consists of four main steps for updating model parameters. In each round, clients initially obtain the global model from the server, then update their local model through gradient descent using their own private data. Next, clients send their updated local model to the server, which aggregates them to create a new global model for the next round.


Many studies have attempted to improve FedAvg to better handle Non-IID data. However, most of these studies focus on distribution bias resulting from either class imbalance or sample size imbalance~\cite{Durmus,Li,Tan}. Yet, the model's classification accuracy and convergence stability can be severely impacted when private data is distributed across multiple domains, such as with feature shifts~\cite{Kairouz} in autonomous driving where different environmental distributions (e.g. weather) cause client data to differ from that of other clients. However, the issue is often more nuanced, with label shifts~\cite{Kairouz} also occurring in widely distributed private data.


To address Non-IID in Federated Learning, Li et al.~\cite{Xiaoxiao} introduced a normalizing layer to the local model, while Luo et al.~\cite{Zhengquan} proposed Disentangled Federated Learning, which separates cross-invariant and domain-specific attributes into two complementary branches. However, these methods have limitations in accounting for local model heterogeneity and may involve a large number of parameters in the communication process between the server and clients.

\subsection{Privacy Preserving}
\subsubsection{Differential Privacy}

Differential privacy is a mathematical definition of privacy that can be used to prove that published data satisfies a certain private property. It is a property of algorithms, not data. For communication based on gradient space, Zhu et al.~\cite{Zhu} proposed a method to intercept gradient information and reconstruct the training data. Differential privacy limits the influence of an individual and reduces the attacker's inference ability~\cite{Abadi}. The formal definition~\cite{Dwork} for differential privacy is defined as
\begin{definition} 
For the adjacent datasets $D$ and $D^\prime$, all possible outputs are $O$, and the mechanism $F$ satisfies
\end{definition}

\begin{equation}
\frac{Pr[F(D) \in O]}{Pr[F(D^\prime) \in O]}\le e^\epsilon
\end{equation}

To satisfy differential privacy, noise is added to the output of the algorithm $f$. This noise is proportional to the sensitivity of the output, where sensitivity measures the maximum change in the output due to the inclusion of a single instance of data. The sensitivity $S_f$ ~\cite{Dwork} of Algorithm $f$ is defined as

\begin{equation}
S_f=\max_{D,D^\prime:d(D,D^\prime)\le 1} |f(D)-f(D^\prime)|
\end{equation}

where $d(D,D^\prime)$ represents the distance between two datasets $D$ and $D^\prime$.

One of the mechanisms to achieve differential privacy is the Gaussian mechanism. The Gaussian mechanism ~\cite{Cynthia} is defined as

\begin{equation}
F(D)=f(D)+N(0,S_f^2\sigma^2)
\end{equation}

where $N(0,S_f^2\sigma^2)$ is the gaussian distribution with mean $0$ and standard deviation $S_f\sigma$. The gaussian mechanism to function $f$ of sensitivity $S_f$ satisfies $(\varepsilon ,\delta )$-differential privacy if $\varepsilon$ and $\delta$ satisfies certain conditions~\cite{Cynthia}.

\subsubsection{Homomorphic Encryption}
It guarantees the following properties
\begin{equation}
Enc(m_1)\circ Enc(m_1) = Enc(m_1+m_2)
\end{equation}

where $\circ$ means "composition" of functions. The scheme is used for privacy protection. Because an untrusted server can perform operations directly on encrypted values.
The additive homomorphic scheme is the Paillier cryptosystem~\cite{Paillier}. 

Damgård et al.~\cite{Dam} proposes a threshold variant of the Paillier cryptosystem, which allows a group of clients to share a key while ensuring that any subset of clients smaller than a predefined threshold can not decrypt the data.

\section{FedPH}
\subsection{Problem Formulation}
In this section, we begin with the Federated Learning Framework in general, characterize the issue, and describe the global objective.

\subsubsection{General Federated Learning Framework} 
According to FedAvg, the global objective of the general Federated Learning Framework for $m$ clients is
\begin{equation}
\min_{w_1,w_2,...,w_m} \frac{1}{m} \sum_{i=1}^{m} \frac{|D_i|}{N}L_i(w_i;D_i)
\end{equation}
where $w_i$ is the local model parameters for the $i$-th client and finally $w_1=w_2=...=w_m$; $m$ is the number of clients; $L_i$ is the local loss function for the $i$-th client; $D_i$ is the local dataset for the $i$-th client; $|D_i|$ is the sample size of the local dataset for the $i$-th client; $N$ is the total number of samples for all clients. 

The local models in $w_1=w_2=...=w_m$ are assumed to be isomorphic, which implies that the clients' computational capabilities are equivalent. If the data distributed in clients are heterogeneous, the local models could not perform well. We suggest a novel Federated Learning approach to address the above issues.

\subsubsection{Proposed Federated Learning Framework} 
\begin{figure}
\includegraphics[width=\textwidth]{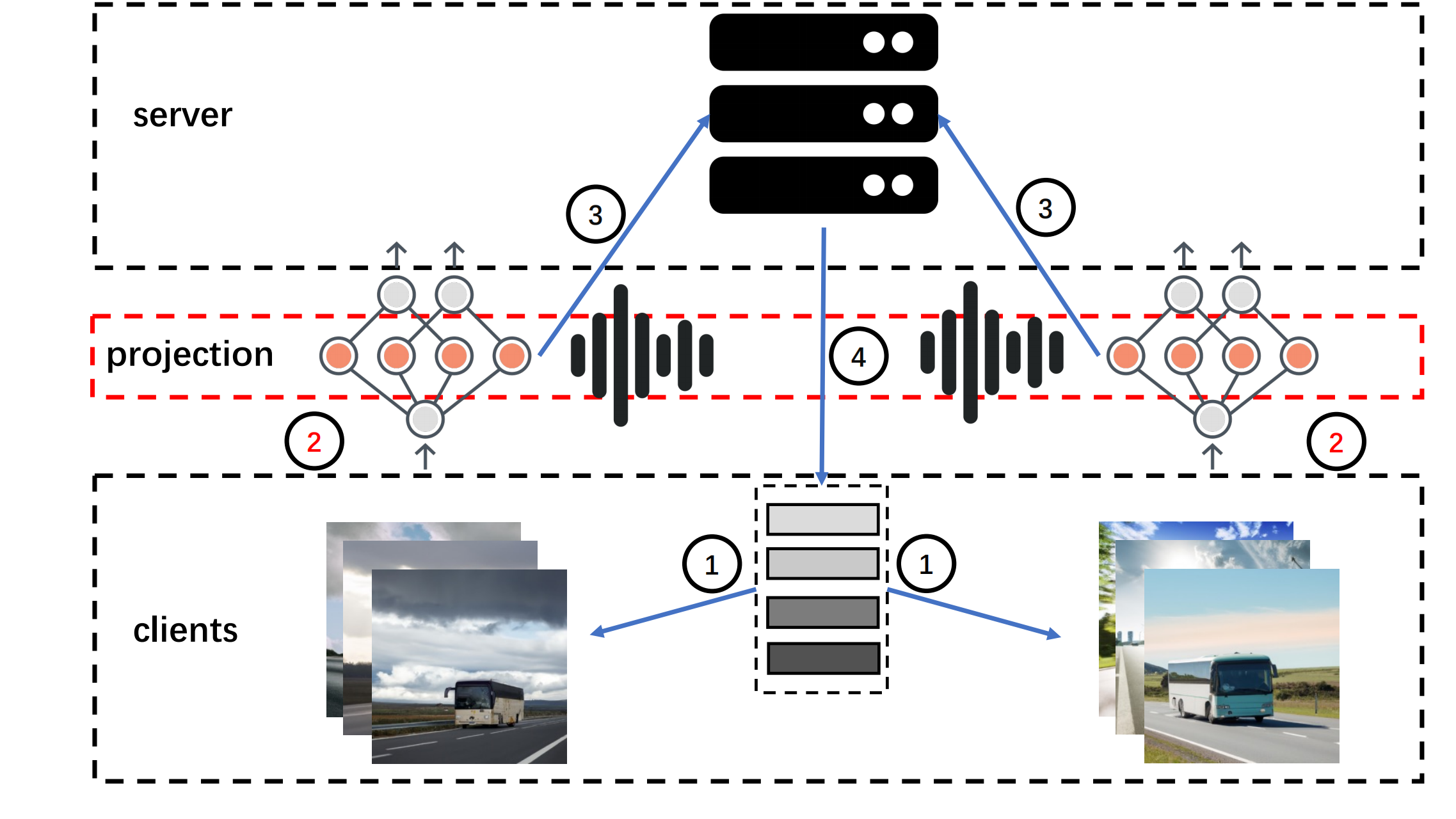}
\caption{(1) The server distributes the global embedding vectors to the clients. (2) Using their private data and the global embedding vectors, clients update their local model and local embedding vectors. (3) Clients send their local embedding vectors back to the server. (4) The server updates the global embedding vectors using the received local embedding vectors.} \label{fig1}
\end{figure}

We suggested Federated Learning method permits $w_1 \neq w_2 \neq ... \neq w_m$, which is different from most Federated Learning methods. FedPH is shown in Figure ~\ref{fig1}.

The pre-trained backbone is fixed for the $i$-th client, and the local dataset $D_i$ is not shared. At least two components make up the local model. (1) Encoder $r(\cdot;\phi^*):x^d\to x^{d_a}$, The $i$-client inputs the raw data $x^d$ to the fixed backbone, and gets the feature vector $x^{d_a}$, which maps the raw data $x$ of size $d$ to a feature vector of size $d_a$. (2) Projection $h(\cdot;\theta _i):x^{d_a}\to x^{d_b}$, The $i$-client inputs $x^{d_a}$ to the unfixed network and gets $x^{d_b}$, which is the mapping process for the embedding space.

\begin{definition}
$r$ represents the embedding function of the backbone.
$x$ represents a sample from the local dataset.
$\phi^*$ represents the parameters of the pre-trained backbone.
To map the backbone output to another embedding space for $i$-client, the projection network $h$ parameterized by $\theta _i$ is used.
The output of the projection network is computed as
\end{definition}

\begin{equation}
z(x)=h(r(x,\phi^*);\theta _i)
\end{equation}

\subsection{Method}
We propose to share embedding vectors between the server and clients to improve the performance of local models. Compared to sharing information through the gradient space, sharing through embedding vectors has several advantages: (1) it requires fewer parameters than sharing models, making it more computationally and communicationally efficient for privacy protection, (2) it uses the embedding vectors as regularization parameters, reducing the impact of data heterogeneity on local model accuracy, and (3) it does not require isomorphic local models as the embedding vectors are used for aggregation.

\subsubsection{Local Embedding Vectors}
We decide to use the embedding vectors as information carriers to extract features from private data.
The mean of sample projections from the same class $j$ serves as the representative for the embedding vectors $C_i^j$ for the $i$-th client.

\begin{equation}
C_i^j=\frac{1}{|D_{i,j}|}\sum_{(x,y)\in D_{i,j}}{z(x)}
\end{equation}

where $C_i^j$ denotes the $j$-class embedding vector of the $i$-th client; $D_i^j$ denotes the $j$-class samples of the $i$-th client. The local embedding vectors are transferred to the server for information aggregation when the $i$-th client has finished the calculation locally.

\subsubsection{Global Embedding Vectors} After receiving the local embedding vector sets $\{C_i\}^m_{i=1}$, the server calculates the global prototype as

\begin{equation}
\overline{C}^j=\frac{1}{|N_{j}|}\sum_{i=1}^{m}{\frac{|D_{i,j}|}{N_{j}}\cdot C_i^j}
\end{equation}

where $N_j$ denotes the set of the $j$-class samples among all clients. $|N_j|$ denotes the number of $N_j$. The global embedding vector set denotes as $\overline{C} = \left \{ \overline{C}^1,\overline{C}^2...  \right \}$.
Through the server, the global embedding vector aggregates the information from the local embedding vectors.

\begin{equation}
N_{j}=\sum_{i=1}^{m} D_i^j
\end{equation}

\subsubsection{Reducing Noise with THE}
\begin{figure}
\includegraphics[width=\textwidth]{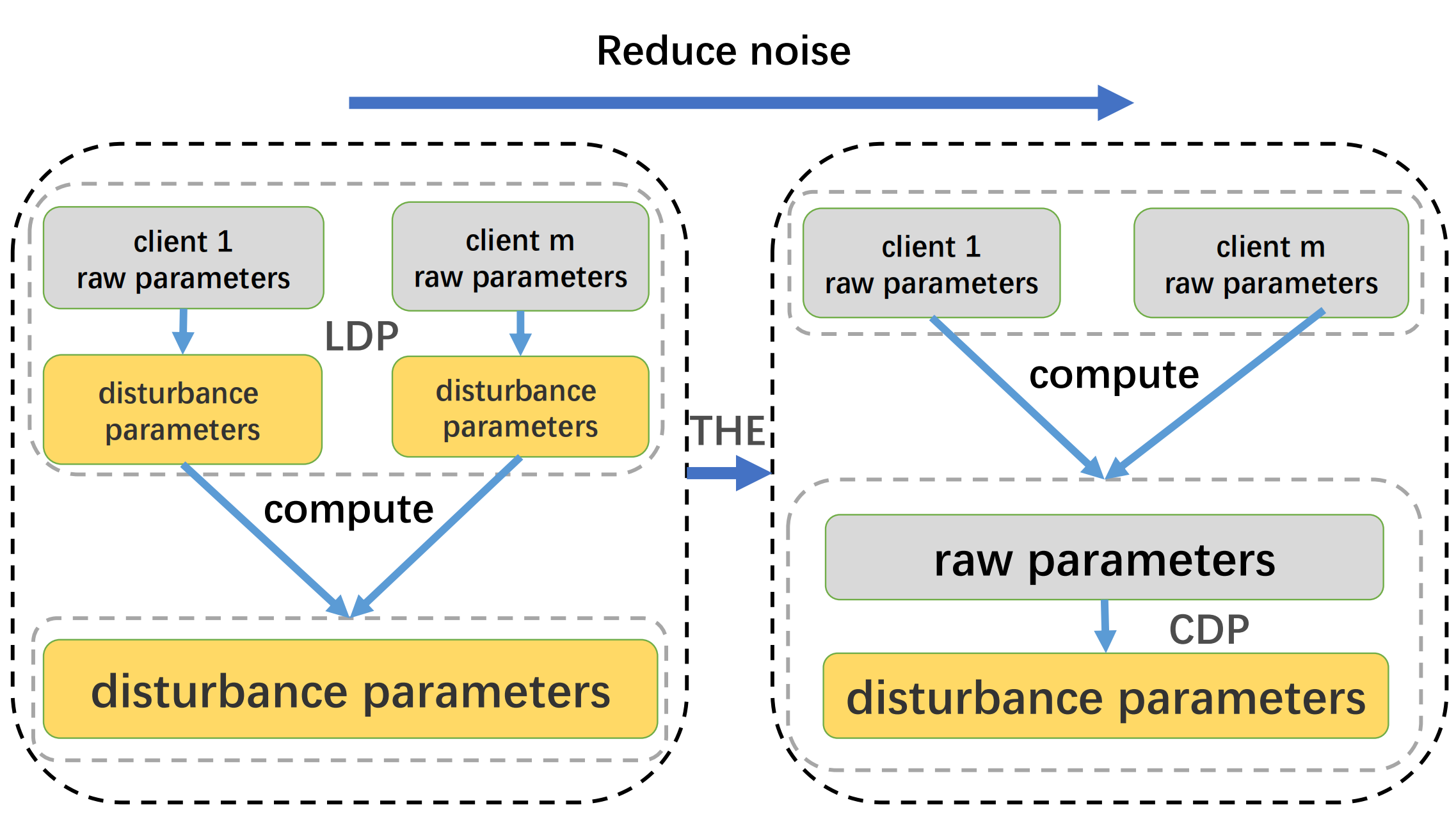}
\caption{Compared to centralized differential privacy (CDP), local differential privacy (LDP) typically requires higher noise levels to achieve the same level of privacy protection. To mitigate this issue, we propose a threshold homomorphic encryption (THE) approach that enables LDP with reduced noise levels.} \label{fig2}
\end{figure}

The threshold homomorphic encryption (THE) algorithm plays a crucial role in the privacy-preserving hybrid method for noise reduction, as depicted in Figure~\ref{fig2}.

\begin{lemma}
$f(D) + N(0,S_f^2\sigma^2)$ satisfies $(\varepsilon ,\delta )$-differential privacy.
\end{lemma}
where the normal distribution $N(0,S_f^2\sigma^2)$ has a mean of $0$ and a standard deviation of $S_f\sigma$.
\begin{proof}
Each client is encrypted using THE proposed in ~\cite{Dam}. $t$ specifies the minimum number of honesty. The threshold is set to $\overline{t} = m - t + 1$, and the noise can be reduced by $t - 1$ times. Each client can return $Enc(C_i^j+N(0,S_f^2\frac{\sigma ^2}{t-1}))$, instead of returning $Enc(C_i^j+N(0,S_f^2\sigma ^2))$. The server first aggregates and then decrypts. The result is $\sum_{i=1}^{m}C_i^j+Y^j$ where $Y^j=N(0,S_f^2\frac{m\sigma ^2}{t-1})$. Since $t - 1 < m$, the noise in the decrypted value is larger than needed to satisfy differential privacy. In addition, THE scheme guarantees that it can not be decrypted even if the maximum number of colluders is $\overline{t}$.
\end{proof}

\section{Local Objective}
\begin{figure}
\includegraphics[width=\textwidth]{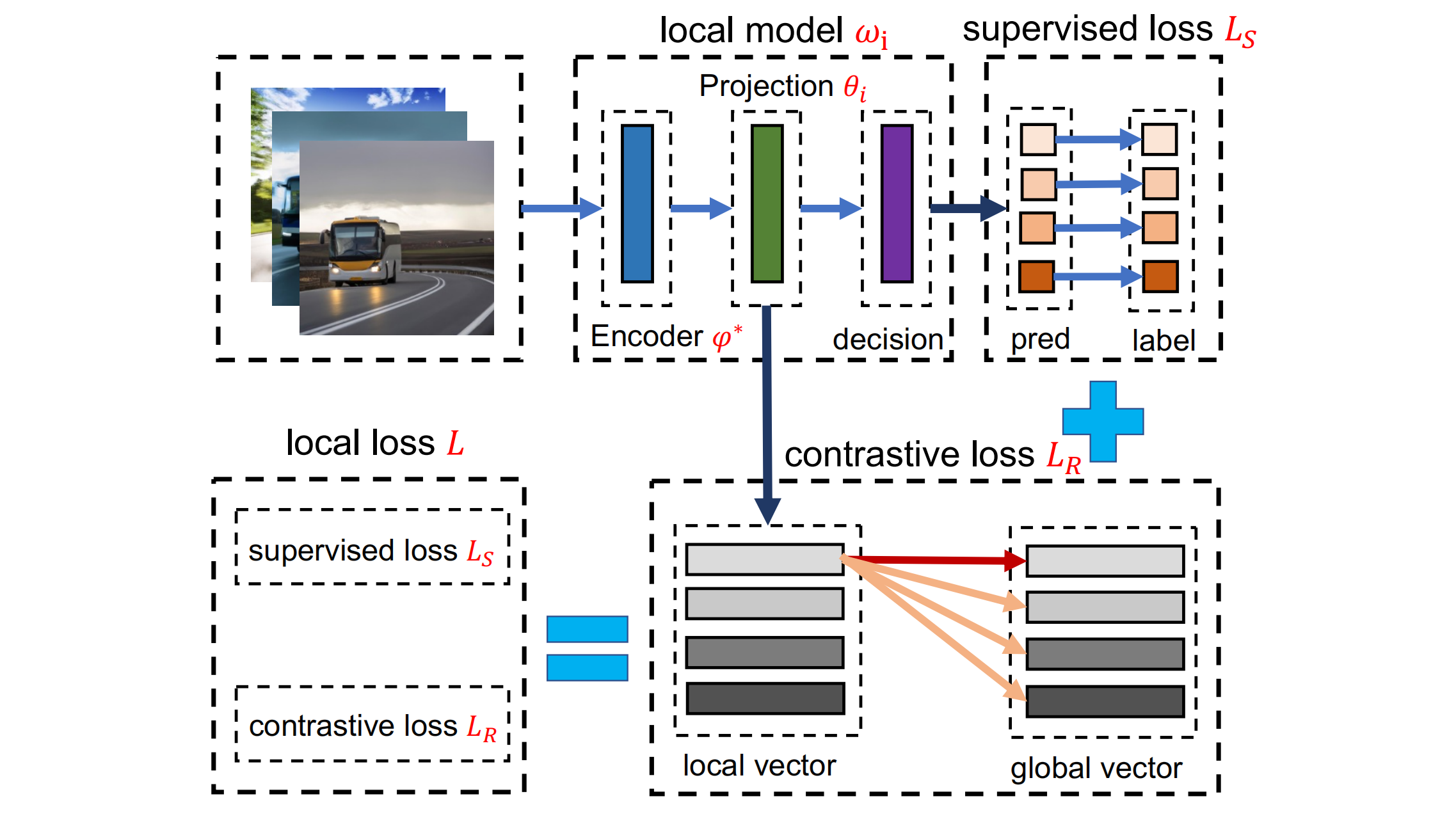}
\caption{The local loss} \label{fig3}
\end{figure}

The local loss is composed of two parts, as illustrated in Figure ~\ref{fig3}. The first part is the cross-entropy loss used in supervised learning, denoted by $L_S$. The second part is the contrastive loss of the embedding vectors, denoted by $L_R$.

Suppose the $i$-client is executing the local training. 
During local training, the $i$-client receives the global embedding vectors from the server and updates the local model as well as the local embedding vectors. 
We extract the embedding vectors from the raw sample $x$ according to the local model ($C^y=z(x)=h(r(x,\phi^*);\theta _i)$). 
Since the global embedding vectors can be better represented, our goal is to reduce the distance between $C^y$ and $\overline{C}^j$ $(y=j)$ and increase the distance between $C^y$ and $\overline{C}^j$ $(y\ne j)$. Similar to the NT-Xent loss~\cite{Sohn}, we define the contrastive loss of embedding vectors as

\begin{equation}
L_R=-log(\frac{exp(dis(C^y,C^j)/t)}{exp(dis(C^y,C^j)/t)+\sum_{y\ne j} exp(dis(C^y,C^j)/t)})
\end{equation}

where $t$ denotes a temperature parameter. The measurement distance function can be $L_1$, $L_2$, and cosine. The loss of a batch $(x,y)$ is computed by

\begin{equation}
L=L_S(\omega _i;(x,y))+\lambda \cdot L_R(\phi^*;\theta _i;\overline{C};(x,y))
\end{equation}

where $\lambda$ is a hyper-parameter to control the weight of embedding vector contrastive loss. The local objective is to minimize
\begin{equation}
\min E_{(x,y)\sim D_i}[L_S(\omega _i;(x,y))+\lambda \cdot L_R(\phi^*;\theta _i;\overline{C};(x,y))]
\end{equation}


Algorithm 1 outlines our proposed Federated Learning approach. During local training, clients utilize stochastic gradient descent to update their personalized local model and local embedding vectors using private data, with the objective function defined in Eq.(12). At each round, the server sends the global embedding vectors to clients and updates them via a weighted average.
\begin{algorithm}
\setcounter{algorithm}{0}
\caption{FedPH}
\label{alg:Framwork}

\renewcommand{\algorithmicensure}{\underline{\textbf{Input:}}} 
\begin{algorithmic}[1] 
\ENSURE ~~\\ 
    number of communication rounds $L$,
    number of clients $m$,
    number of local epochs $E$,
    global embedding vectors $\overline C$, local embedding vectors $C$,
    the minimum number of honesty $t$,
    the maximum number of colluders $\overline{t}$,
    randomly selecting $\overline{t}$ clients from $m$ clients to form a set $P$
\end{algorithmic}

\renewcommand{\algorithmicensure}
{\underline{\textbf{Output:}}} 
\begin{algorithmic}[1] 
\ENSURE ~~\\ 
    The final global embedding vectors $\overline {C}^{L}$
\end{algorithmic}

\renewcommand{\algorithmicensure}{\underline{\textbf{Server executes:}}} 
\begin{algorithmic}[1] 
\ENSURE ~~\\ 
\STATE Initialize the global embedding vectors $\overline {C}^1$ 

\FOR{$l = 1,2,...,L$}

    \FOR{$i = 1,2,...,m$}
        \STATE $r_i\longleftarrow$ LocalTraining(${i,\overline{C}^{l}}$) 
    \ENDFOR

    \STATE Aggregate local embedding vectors by $r = r_1 \circ r_2 \circ ... \circ r_m$

    \FOR{$i \in P$}
        \STATE $r=Dec_{sk_i}(r)$
    \ENDFOR
    \STATE Update global embedding vectors by $\bar{C}^{l+1} = \frac{r}{m}$
    

\ENDFOR
\end{algorithmic}


\renewcommand{\algorithmicensure}{\underline{\textbf{LocalTraining($i$,$\overline{C}^l$)}}} 
\begin{algorithmic}[1] 
\ENSURE ~~\\ 
\FOR{epoch $i=1,2,...,E$}
    \FOR{each batch $(x,y) \in D_i$}
        \STATE Compute local embedding vectors by Eq.7
        \STATE Compute loss by Eq.11 using local embedding vectors
        \STATE Update local model parameters according to the loss
    \ENDFOR
\ENDFOR
\STATE \textbf{return} $Enc_{pk}(C_i + N(0,S_f^2\frac{\sigma^2}{t-1}))$
\end{algorithmic}
\end{algorithm}
\section{Experiments}
\subsection{Experimental Setup}

We compare FedPH to three other Federated Learning methods: FedAvg~\cite{McMahan}, FedProx~\cite{Li}, and FedProto~\cite{Tan}. We also establish a baseline method, SOLO, in which clients are trained on private data without using Federated Learning.

\begin{figure}
\includegraphics[width=\textwidth]{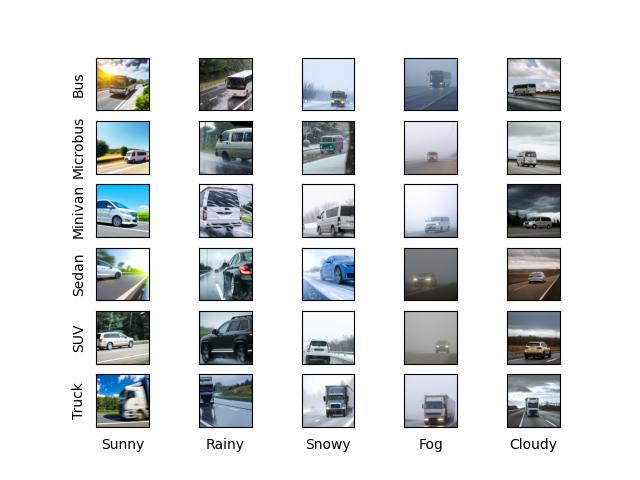}
\caption{The horizontal axis represents different weather classes, with each client corresponding to a specific class. The vertical axis represents vehicle classes, which correspond to private data classes.} \label{fig4}
\end{figure}
\begin{figure}
\includegraphics[width=\textwidth]{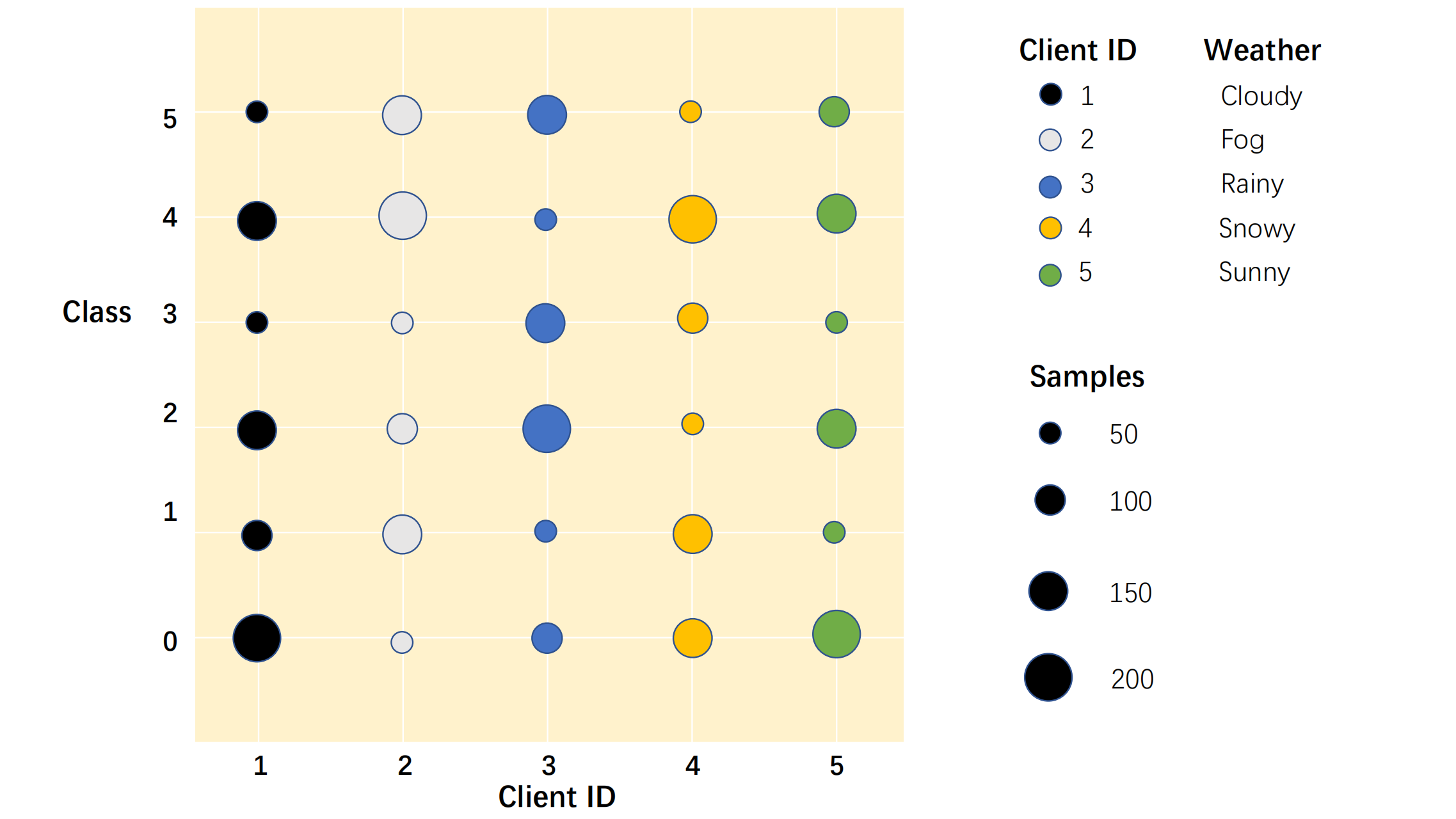}
\caption{Label and feature distributions of the private data vary among clients.} \label{fig5}
\end{figure}

Our experiments were performed on a custom vehicle dataset consisting of 5,000 images that depicted six different types of vehicles and five different weather conditions, as illustrated in Figure~\ref{fig4}. As the weather conditions varied, there were feature shifts observed in the data. We generated label shifts among clients by using the Dirichlet distribution. While there were many Non-IID classes in our dataset, both feature and label shifts are common occurrences in real-world scenarios, as depicted in Figure~\ref{fig5}.

We employ a fully connected layer as the projection head, another fully connected layer as the decision component, and use the pre-trained ResNet-18~\cite{He} as the encoder. It is worth noting that all baselines also adopt the network architecture of FedPH.


We use PyTorch to implement FedPH and the other baseline methods. For all approaches, we adopt the SGD optimizer with a learning rate of 0.001, SGD momentum of 0.5, and SGD weight decay of 0.0001. The batch size is set to 32, and a pre-trained network serves as the backbone for all methods. For the contrastive loss of FedPH, we measure the distance between the local and global embedding vectors using cosine distance and set the temperature parameter to 1.

\subsection{Accuracy}

\begin{figure}
\includegraphics[width=\textwidth]{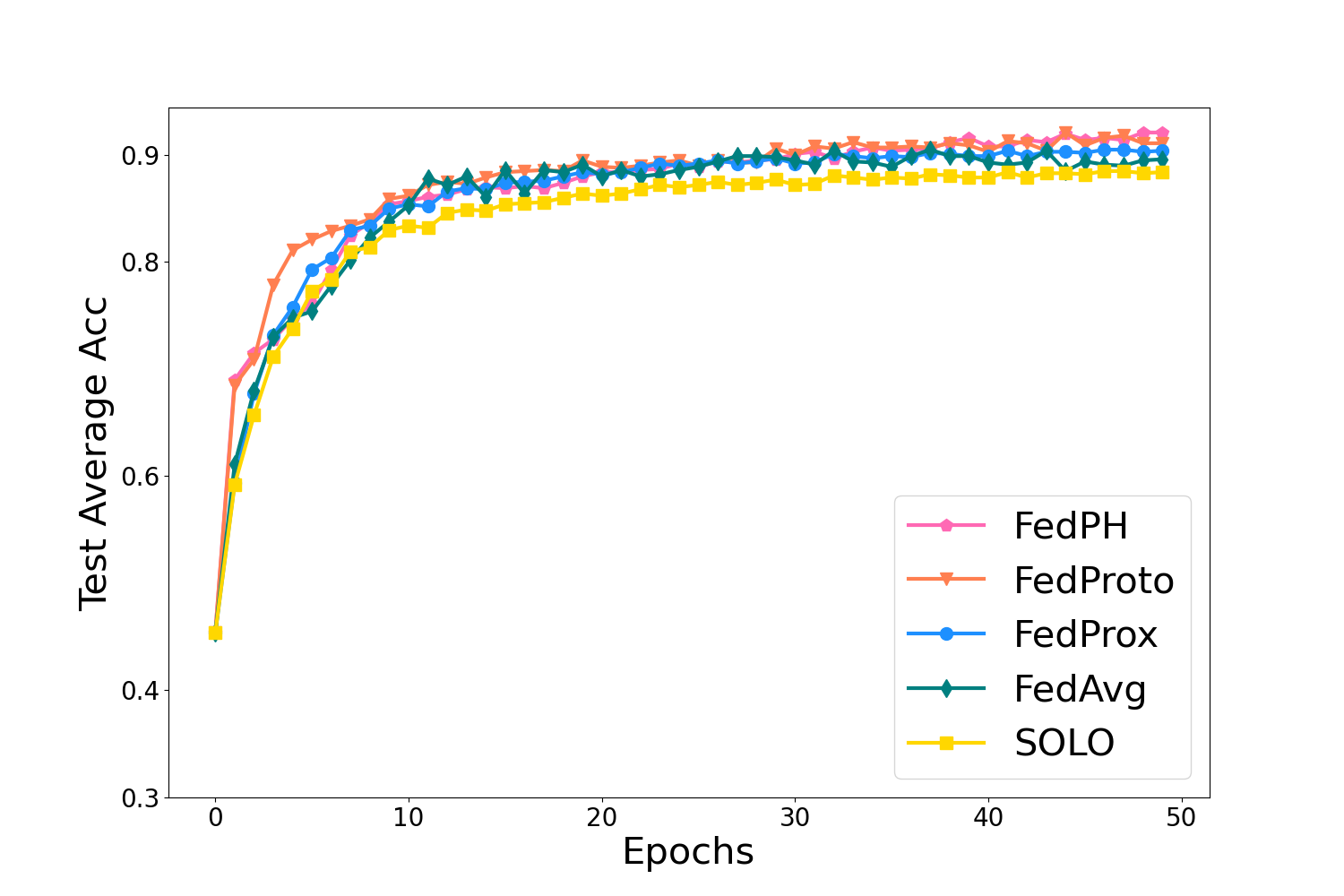}
\caption{} \label{fig6}
\end{figure}

In the vehicle dataset with a Non-IID setting, Federated Learning methods have shown better accuracy than SOLO, as demonstrated in Figure~\ref{fig6}. Among the compared methods, FedPH has demonstrated the best performance, outperforming FedAvg by an average of 2.5\% on supervised learning tasks. Although the precision of FedProto is comparable to that of FedPH, the introduction of contrastive loss results in our suggested FedPH surpassing FedProto by an average of 1\%, as presented in Table~\ref{tab1}. This suggests that FedPH is effective in mitigating the negative effects of Non-IID.

\begin{table}[!htb]
  \centering
  \caption{Comparison of top-1 accuracy}
  \label{tab1}
  \begin{tabular}{|l|l|l|}
  \hline
    Method     & 5 clients\\
  \hline
    SOLO      & 88.4\% $\pm$ 1.61\%\\
    FedAvg     & 89.6\% $\pm$ 1.78\%\\
    FedProx     & 90.4\% $\pm$ 1.52\%\\
    FedProto  & 91.1\% $\pm$ 0.37\%\\
    FedPH  & \textbf{92.1\% $\pm$ 0.24\%}\\
  \hline
  \end{tabular}
\end{table}

In FedPH, the embedding vector is a shared parameter between the server and clients that effectively captures feature representations of high-dimensional data, removing irrelevant information. By incorporating contrastive loss as a regular term in the local loss function, the embedding vectors of similar data are further shortened, resulting in significant performance gains in the decision layer of the local model. As a result, FedPH achieves superior results.

\subsection{Communication Efficiency}
Due to the limitations of the current communication infrastructure, Federated Learning encounters significant challenges related to communication costs. Therefore, we monitored the size of the parameters for each round of communication.

\begin{table}[!htb]
  \centering
  \caption{Comparison of parameter size}
  \label{tab2}
  \begin{tabular}{|l|l|l|}
    \hline
    Method     & Params\\
    \hline
    FedAvg     & 33200\\
    FedProx     & 33200\\
    FedProto  & \textbf{384}\\
    FedPH  & \textbf{384}\\
  \hline
  \end{tabular}
\end{table}

Table~\ref{tab2} shows that FedPH has significantly fewer parameters than other methods, and is much more efficient in terms of communication. This suggests that when there is high model heterogeneity, sharing more parameters does not necessarily lead to better outcomes. Thus, it is important to determine which components should be shared in order to optimize the current system.

\subsection{Model Heterogeneity}
In the configuration with model heterogeneity, small variations in model structure between clients are considered, with some having 2 or 3 fully connected layers. Due to differing model parameters, it becomes challenging to average the parameters.

\begin{figure}
\includegraphics[width=\textwidth]{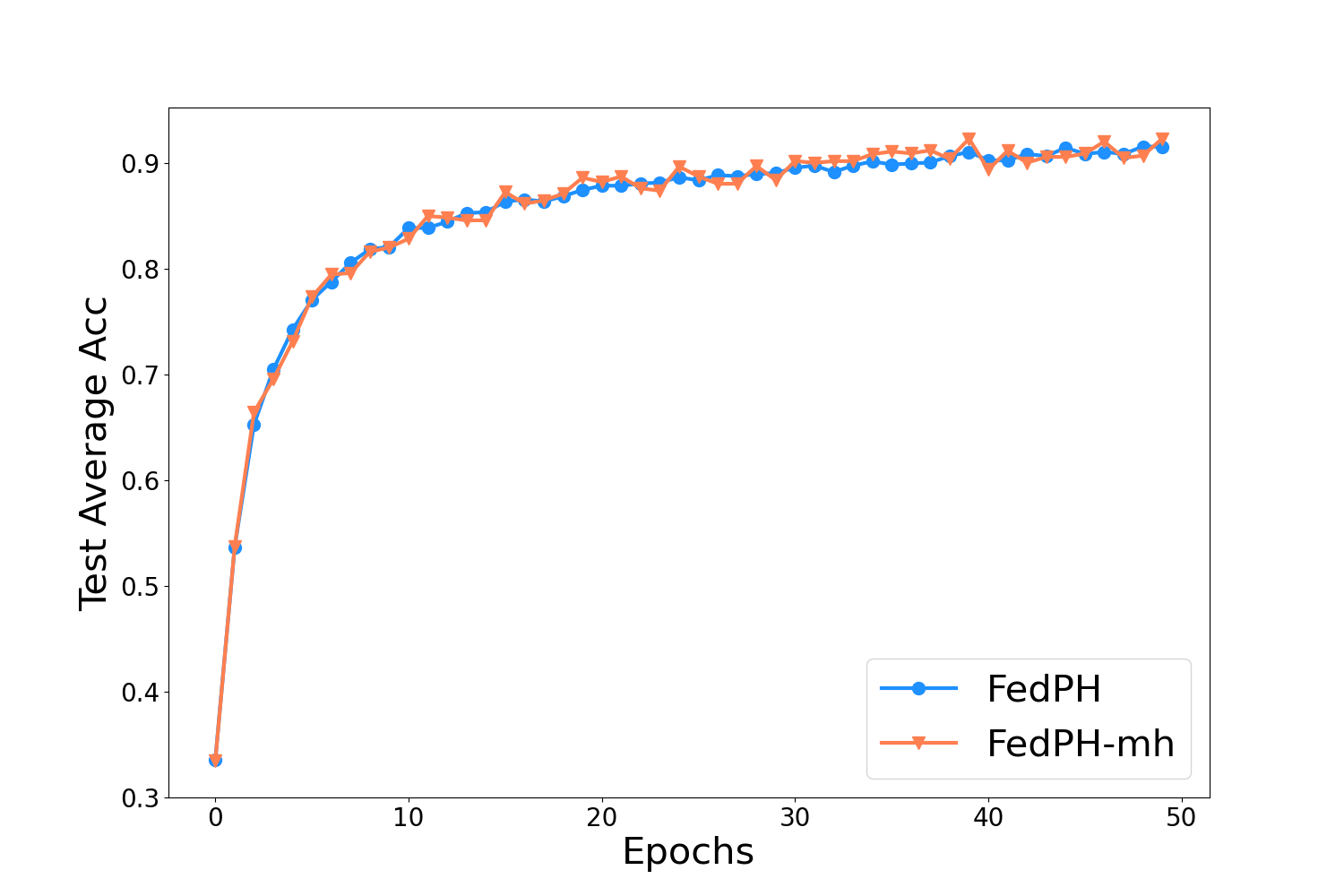}
\caption{FedPH-mh is a variation of FedPH that specifically addresses the challenge of local model heterogeneity.} \label{fig7}
\end{figure}

Figure~\ref{fig7} illustrates how FedPH can achieve consistency among different clients. Unlike traditional Federated Learning methods that rely on model averaging, FedPH utilizes a more personalized approach to better fit private data in terms of both value and shape of model parameters. By abandoning model averaging, FedPH avoids potential issues related to model heterogeneity and achieves greater robustness.
\subsection{Privacy-preserving}
To track changes in Federated Learning's performance, we integrate it with the privacy-preserving method.
More specifically, we perturb the local embedding vectors by adding noise with Gaussian distribution.
We make sure that $(\epsilon,\delta)$-differential privacy is satisfied by the aggregated embedding vectors.

\begin{figure}
\includegraphics[width=\textwidth]{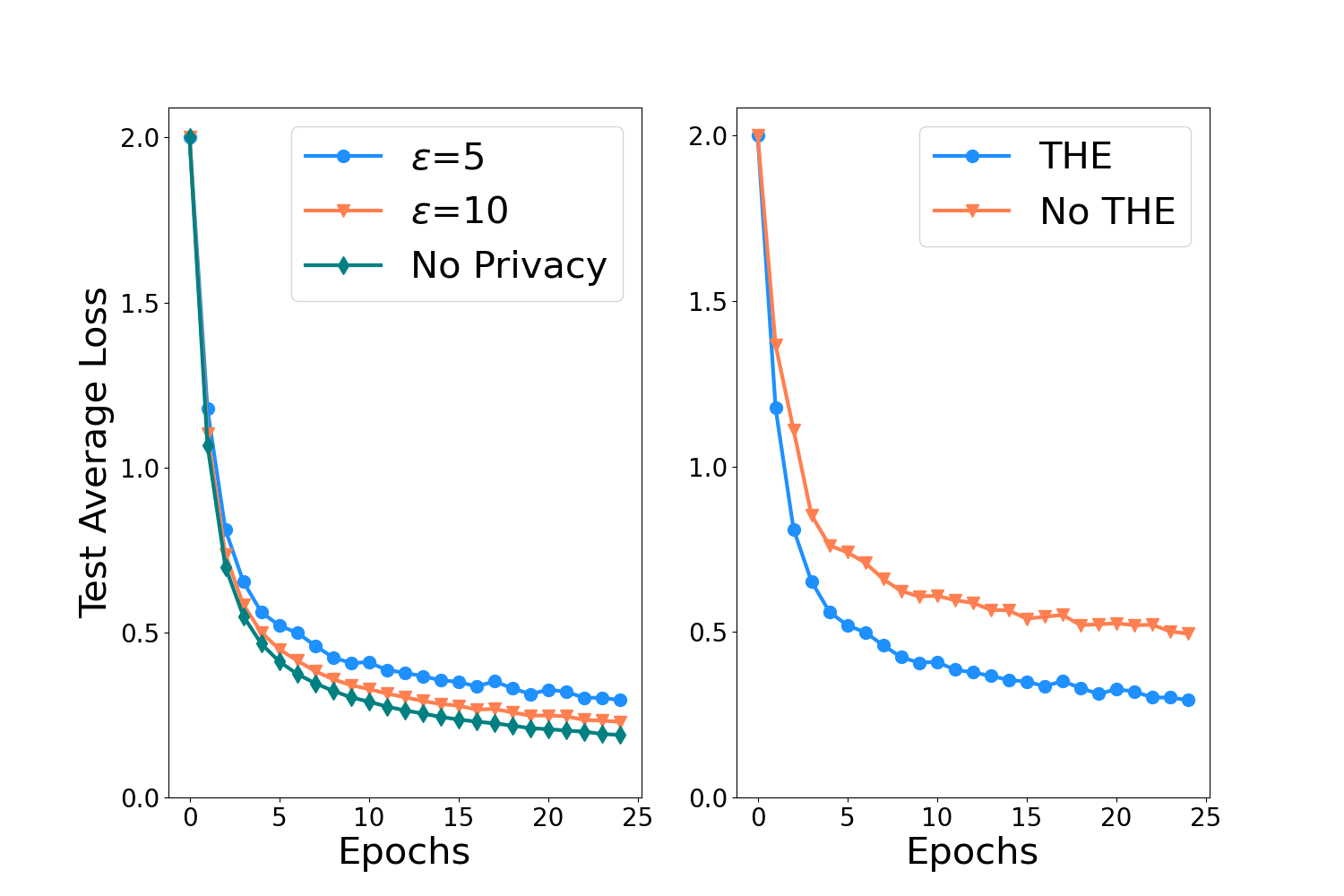}
\caption{Comparing the right half of figure with $\epsilon=5$} \label{fig8}
\end{figure}

In this experiment, the threshold was set to 3 and $\delta$ was set to $10e-5$.
Relaxing the privacy guarantees(increasing $\epsilon$) reduces the associated loss, as shown in the left half of Figure~\ref{fig8}. Applying the threshold homomorphic encryption approach reduces the impact of the noise required to satisfy differential privacy on the model, as shown in the right half of Figure ~\ref{fig8}.
\begin{table}[!htb]
  \centering
  \caption{Comparison of time consumption}
  \label{tab3}
  \begin{tabular}{|l|l|l|}
    \hline
    parameter     & time\\
    \hline
    model parameters& 3.372$\pm$0.0159s\\
    embedding vectors & \textbf{0.039$\pm$0.004s}\\
  \hline
  \end{tabular}
\end{table}

According to Table~\ref{tab3}, we find that selecting embedding vectors as aggregate parameters in privacy preservation is faster than selecting model parameters. It is important to note that this table only records the encryption process for one communication round. However, this advantage will be further amplified in multiple communication rounds.

In conclusion, FedPH integrates a privacy-preserving method that effectively protects privacy without visibly impacting performance and conserves computing resources.

\section{Conclusion}
In this study, we propose a novel Federated Learning method that combines differential privacy and threshold homomorphic encryption to protect local data privacy while minimizing the impact on local model accuracy, ensuring both privacy and security. Our approach achieves excellent privacy protection and accurate prediction results in heterogeneous contexts.
Unlike traditional approaches that share information based on the gradient space, our approach shares embedding vectors between the server and clients. We conduct experiments to demonstrate the effectiveness of our method.
%
%
%
%

\end{document}